\documentclass{article}
\usepackage{microtype}
\usepackage{graphicx}
\usepackage{subcaption}
\usepackage{mathtools}
\usepackage{amsthm}

\usepackage{amsmath, amssymb}
\usepackage{mathabx}
\usepackage{xspace}
\usepackage{hyperref}

\usepackage[preprint]{icml2026}
\usepackage{booktabs}

\usepackage[capitalize,noabbrev]{cleveref}

\theoremstyle{plain}

\theoremstyle{definition}

\theoremstyle{remark}

\usepackage[textsize=tiny]{todonotes}
\def \alambic {\includegraphics[width=0.03\linewidth]{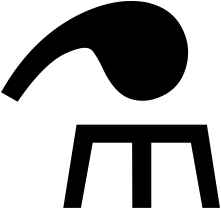}\xspace}
\def \arrowtrans {\includegraphics[width=0.03\linewidth]{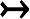}\xspace}
\icmltitlerunning{Knowledge Diversion for Efficient Morphology Control and Policy Transfer}

\begin{document}

\twocolumn[
  \icmltitle{Knowledge Diversion for Efficient Morphology Control and Policy Transfer}



  \icmlsetsymbol{equal}{*}

  \begin{icmlauthorlist}
    \icmlauthor{Fu Feng}{add1,add2,equal}
    \icmlauthor{Ruixiao Shi}{add1,add2,equal}
    \icmlauthor{Yucheng Xie}{add1,add2}
    \icmlauthor{Jianlu Shen}{add1,add2}
    \icmlauthor{Jing Wang}{add1,add2}
    \icmlauthor{Xin Geng}{add1,add2}
  \end{icmlauthorlist}

  \icmlaffiliation{add1}{School of Computer Science and Engineering, Southeast University, Nanjing, China}
  \icmlaffiliation{add2}{Key Laboratory of New Generation Artificial Intelligence Technology and Its Interdisciplinary Applications (Southeast University), Ministry of Education, China}

  \icmlcorrespondingauthor{Jing Wang}{wangjing91@seu.edu.cn}
  \icmlcorrespondingauthor{Xin Geng}{xgeng@seu.edu.cn}

  \icmlkeywords{Machine Learning, ICML}

  \vskip 0.3in
]



\printAffiliationsAndNotice{\icmlEqualContribution}

\begin{abstract}
  Universal morphology control aims to learn a universal policy that generalizes across heterogeneous agent morphologies, with Transformer-based controllers emerging as a popular choice.
  However, such architectures incur substantial computational costs, resulting in high deployment overhead, and existing methods exhibit limited cross-task generalization, necessitating training from scratch for each new task.
  To this end, we propose \textbf{DivMorph}, a modular training paradigm that leverages knowledge diversion to learn decomposable controllers. 
  DivMorph factorizes randomly initialized Transformer weights into factor units via SVD prior to training and employs dynamic soft gating to modulate these units based on task and morphology embeddings, separating them into shared \textit{learngenes} and morphology- and task-specific \textit{tailors}, thereby achieving knowledge disentanglement.
  By selectively activating relevant components, DivMorph enables scalable and efficient policy deployment while supporting effective policy transfer to novel tasks.
  Extensive experiments demonstrate that DivMorph achieves state-of-the-art performance, achieving a 3$\times$ improvement in sample efficiency over direct finetuning for cross-task transfer and a 17$\times$ reduction in model size for single-agent deployment.
\end{abstract}

\section{Introduction}
Reinforcement learning (RL) is a fundamental approach for robotic control~\cite{ju2022transferring}, yet policies often transfer poorly across heterogeneous morphologies or tasks~\cite{wangprogrammatically, he2024efficient}. 
Agents with different morphologies require distinct policies, and even a single morphology relies on task-specific strategies~\cite{yang2020multi, trabucco2022anymorph}. Consequently, standard RL pipelines train from scratch for each morphology–task pair, lacking adaptive generalization mechanisms~\cite{singh2022reinforcement}.

% d: enabling agents to acquire complex behaviors through direct interaction.

\begin{figure}
    \centering
    \includegraphics[width=\linewidth]{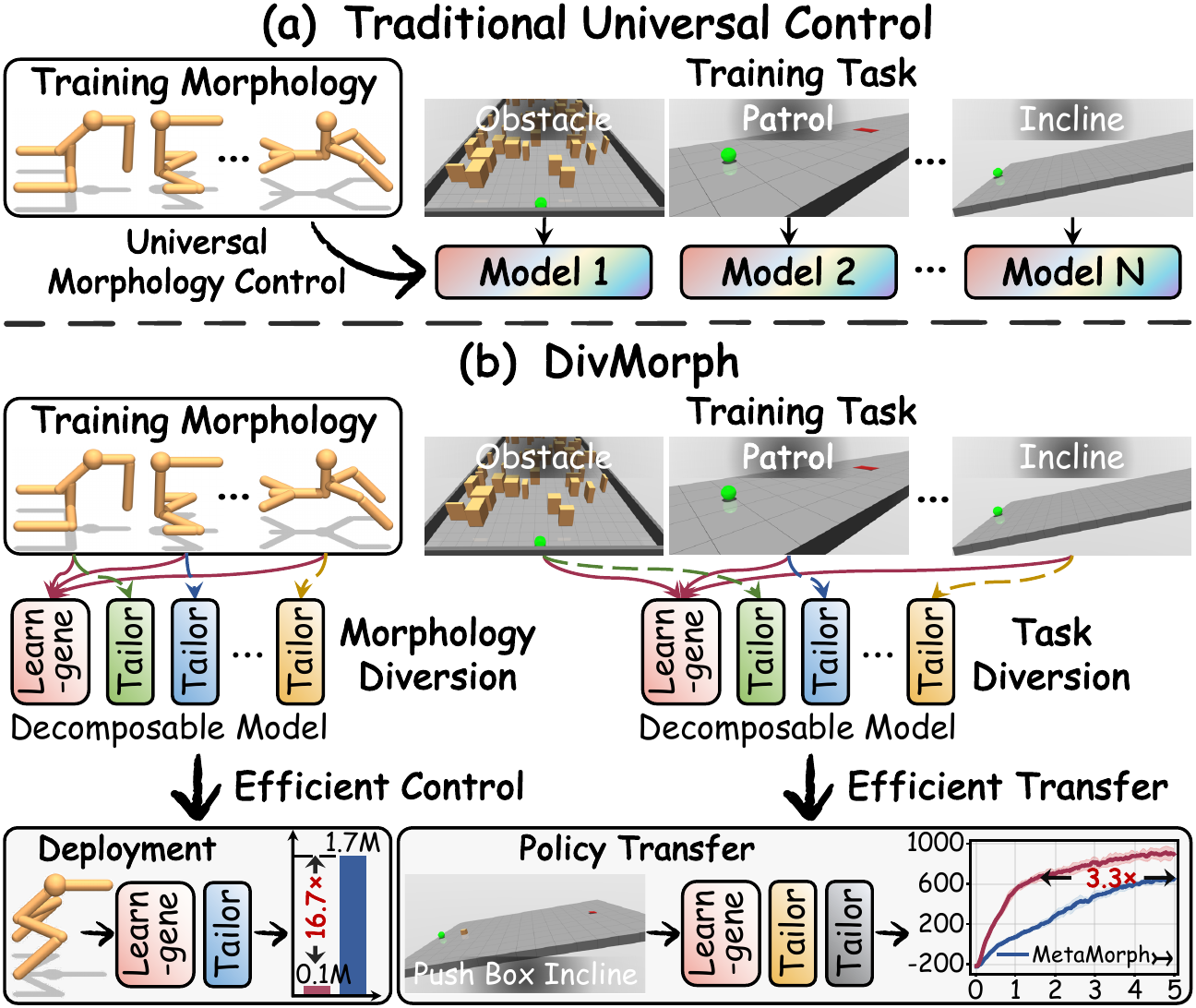}
    \vspace{-0.2in}
    \caption{(a) Traditional universal morphology control trains a single controller for diverse morphologies but remains task-specific, requiring separate training for each new task. 
    (b) DivMorph leverages knowledge diversion to train a modular network that decouples shared, morphology-specific, and task-specific components, enabling adaptive recomposition for universal morphology control with 3$\times$ higher sample efficiency on new tasks, and 17$\times$ smaller deployment models for a given agent morphology.}
    \label{fig:moti}
    \vspace{-0.23in}
\end{figure}

To improve policy reuse, universal morphology control has been proposed to learn a single controller that generalizes across diverse agent morphologies~\cite{wang2018nervenet, kurin2020my, gcnt2025luo}.
Recent work, such as MetaMorph~\cite{guptametamorph}, advances this direction by replacing the MLP policy with a morphology-aware Transformer that represents each agent as a sequence of limbs.
ModuMorph~\cite{xiong2023universal} further exploits morphology context to improve control performance. 
However, Transformer based controllers incur substantial memory and computation overhead, making per agent deployment costly~\cite{hu2024transforming}.
HyperDistill~\cite{xiong2024distilling} mitigates this overhead by using a hypernetwork to transfer Transformer knowledge into a compact MLP, but this compression compromises generalization to unseen morphologies and limits architectural flexibility.

Despite these advances, existing approaches generalize only at the morphology level and remain task-specific (Fig.~\ref{fig:moti}a), resulting in poor policy reuse across tasks, while their structural rigidity incurs additional deployment overhead.
This motivates a central question: \textbf{\textit{can we train a structurally flexible Transformer that enables efficient control across diverse morphologies while supporting effective policy reuse across tasks?}}

Recently, knowledge diversion (KDiv)~\cite{xie2025kind} has emerged as a pretraining paradigm in generative models~\cite{xie2025divcontrol, xuctrlora}, explicitly disentangling task-agnostic and task-specific knowledge by factorizing network weights into shared \textit{learngenes} and task-specific \textit{tailors}, thereby improving modular reuse and cross-task transferability.
Building on this principle, we introduce DivMorph, which brings knowledge diversion into universal morphology control (Fig.~\ref{fig:moti}b). 
By disentangling shared (i.e., morphology- and task-agnostic) knowledge from morphology- and task-specific components during training, DivMorph enables effective cross-task policy transfer for unified morphology control via transferable common knowledge, and efficient deployment for specific agent morphologies with significantly reduced model size.

To enable knowledge decoupling across morphologies and tasks in transformer-based controllers, DivMorph first decomposes each randomly initialized weight matrix $W$ via Singular Value Decomposition (SVD) as $W = U \Sigma V^\top$. 
Each singular vector pair $(u_i, v_i)$ in $U$ and $V$ is treated as a factor unit and assigned to one of three categories: shared \textbf{learngenes}, \textbf{morphology-specific tailors}, or \textbf{task-specific tailors}.
During knowledge diversion, the singular values $\Sigma$ are modulated by morphology- and task-aware gates, which control the recombination of factor units based on task embeddings from textual instructions encoded via a pretrained text encoder and morphology embeddings derived from each agent’s limb joint configuration using a dedicated morphology encoder.
To improve parameter sharing and generalization, DivMorph replaces conventional binary gating~\cite{xie2025kind, xuctrlora} with dynamic soft gates that assign soft weights to tailors in a mixture-of-experts style~\cite{zhou2022mixture}, enabling modular reuse and zero-shot generalization to novel morphologies and tasks.

We conduct extensive experiments in the UNIMAL design space~\cite{gupta2021embodied}, which comprises robots with 15–20 DoFs capable of learning locomotion and mobile manipulation in complex stochastic environments. 
To evaluate both morphology- and task-level generalization, we construct 100 agent morphologies and 5 tasks each for training and testing, following~\cite{gupta2021embodied, guptametamorph}.
Remarkably, DivMorph achieves state-of-the-art performance, offering 3$\times$ higher sample efficiency than direct fine-tuning for new task transfer, while reducing model size by 17$\times$ during deployment for a given agent morphology, comparable to an MLP controller, demonstrating its effectiveness for scalable and flexible policy reuse.

Our contributions are as follows:
1)~We propose DivMorph, the first modular training paradigm for universal morphology control, enabling flexible recomposition for efficient policy deployment.
2)~We introduce a dynamic soft gating mechanism that effectively models tasks and morphologies, enabling enhanced universal control and robust zero-shot policy generalization.
3)~We extend the UNIMAL benchmark to more challenging cross-task policy transfer settings. Extensive experiments demonstrate that DivMorph achieves state-of-the-art performance with substantially improved convergence speed and performance.

% \alambic
% \arrowtrans

\section{Related Work}
\subsection{Universal Morphology Control}
Universal morphology control seeks to learn a single controller capable of operating agents with diverse morphologies, enabling scalable control across large morphology spaces~\cite{nagabandilearning, pathak2019learning, patel2025get}. 
Early MLP-based controllers fail to effectively handle heterogeneous state and action spaces~\cite{ghadirzadeh2021bayesian, feng2023genloco}. 
Subsequent work introduced Graph Neural Networks (GNNs) to better capture morphological topology by enabling structured communication among neighboring actuators~\cite{wang2018nervenet, huang2020one}.
More recently, MetaMorph~\cite{guptametamorph} advanced this direction with a morphology-aware Transformer, and ModuMorph~\cite{xiong2023universal} further leveraged morphology context to improve control performance, though Transformer-based controllers incur substantial computational cost. HyperDistill~\cite{xiong2024distilling} sought to mitigate this cost by distilling the Transformer policy into a compact MLP via a hypernetwork, but the distilled controllers exhibited limited generalization to novel morphologies.
Importantly, existing methods achieve modest morphology-level generalization but remain task-specific, requiring retraining for each new task. 
In contrast, we propose DivMorph, which applies knowledge diversion during training to disentangle morphology-specific and task-specific components, enabling efficient control across morphologies and flexible policy transfer across tasks.

% \vspace{-0.1in}
\subsection{Learngene and Knowledge Diversion}
\textsc{Learngene}~\cite{feng2023genes, wang2023learngene} is a biologically inspired knowledge transfer paradigm that encapsulates task-agnostic knowledge into modular neural units as learngenes for efficient adaptation, substantially improving generalization while mitigating negative transfer~\cite{feng2024transferring}.
Early learngene-based approaches typically condensed task-agnostic knowledge at the layer level. GRL~\cite{feng2023genes} leveraged evolutionary strategies, while Heur-LG~\cite{wang2022learngene} and Auto-LG~\cite{wang2023learngene} relied on heuristics or meta-learning, and other variants~\cite{feng2024wave, xia2024transformer, xie2024fine} imposed structural constraints to regulate parameter sharing.
Although effective, these methods offer limited knowledge disentanglement, limiting adaptability in multi-task and cross-domain scenarios.

Knowledge Diversion (KDiv)~\cite{xie2025kind} advances this line of work by decomposing parameters into task-agnostic learngenes and task-specific tailors, enabling modular reuse through gated routing~\cite{xie2025divcontrol, shi2025fad}.
We extend this paradigm to universal morphology control by disentangling policy networks across both morphology and task dimensions, enabling unified morphology–task control and supporting efficient deployment and effective transfer to novel morphologies and tasks.

\section{Preliminary}
\subsection{Problem Formulation}
We study the problem of learning a \textit{universal policy} that can control a set of $K$ robots with diverse morphologies across $T$ tasks. For a given task $\tau$,
the control process for robot $\kappa$ is formulated as a contextual Markov Decision Process~\cite{hallak2015contextual} $\mathcal{M}_\kappa^{(\tau)} = (\mathcal{S}_\kappa, \mathcal{A}_\kappa, \mathcal{C}_k, T_k, R_{\tau})$, where $\mathcal{S}_\kappa$, $\mathcal{A}_\kappa$, $\mathcal{C}_\kappa$ and $T_\kappa$ denote the state space, action space, morphology context and transition function, while $R_\tau$ denote the task-specific reward, respectively.

We consider robots from a modular design space, where each morphology is represented as a tree of limb nodes with shared local state and action definitions. 
For a robot with $N_\kappa$ limbs, the state and action spaces are $\mathcal{S}_\kappa=\{S_\kappa^{i}\}_{i=1}^{N_\kappa}$ and $\mathcal{A}_\kappa=\{A_\kappa^{i}\}_{i=1}^{N_\kappa}$. 
The task context $\mathcal{C}_\tau$ includes node-specific attributes (e.g., limb size, mass, initial position) and an adjacency matrix encoding the morphology topology.

Let $s_{\kappa,t}$, $a_{\kappa,t}$, and $r_{\kappa,t}$ denote the state, action, and reward of robot $\kappa$ at time $t$. The goal is to learn a universal policy $\pi_\theta(a_{\kappa,t}\mid s_{\kappa,t}, c_\kappa)$ that maximizes the average return across all training morphologies: $\max_{\theta} \; \frac{1}{K} \sum_{\kappa=1}^{K} \sum_{t=0}^{H} r_{\kappa,t}$, where $H$ is the task horizon. 
In addition to maximizing returns on the training morphologies and tasks, we require the universal policy $\pi_\theta$ to efficiently control unseen morphologies $\kappa^\ast \notin \{1,\dots,K\}$ and rapidly adapt to novel tasks $\tau^\ast \notin \{1,\dots,T\}$, enabling zero-shot deployment and cross-task transfer with minimal training.

\subsection{Morphology-Aware Transformer}
\begin{figure*}
    \centering
    \includegraphics[width=\linewidth]{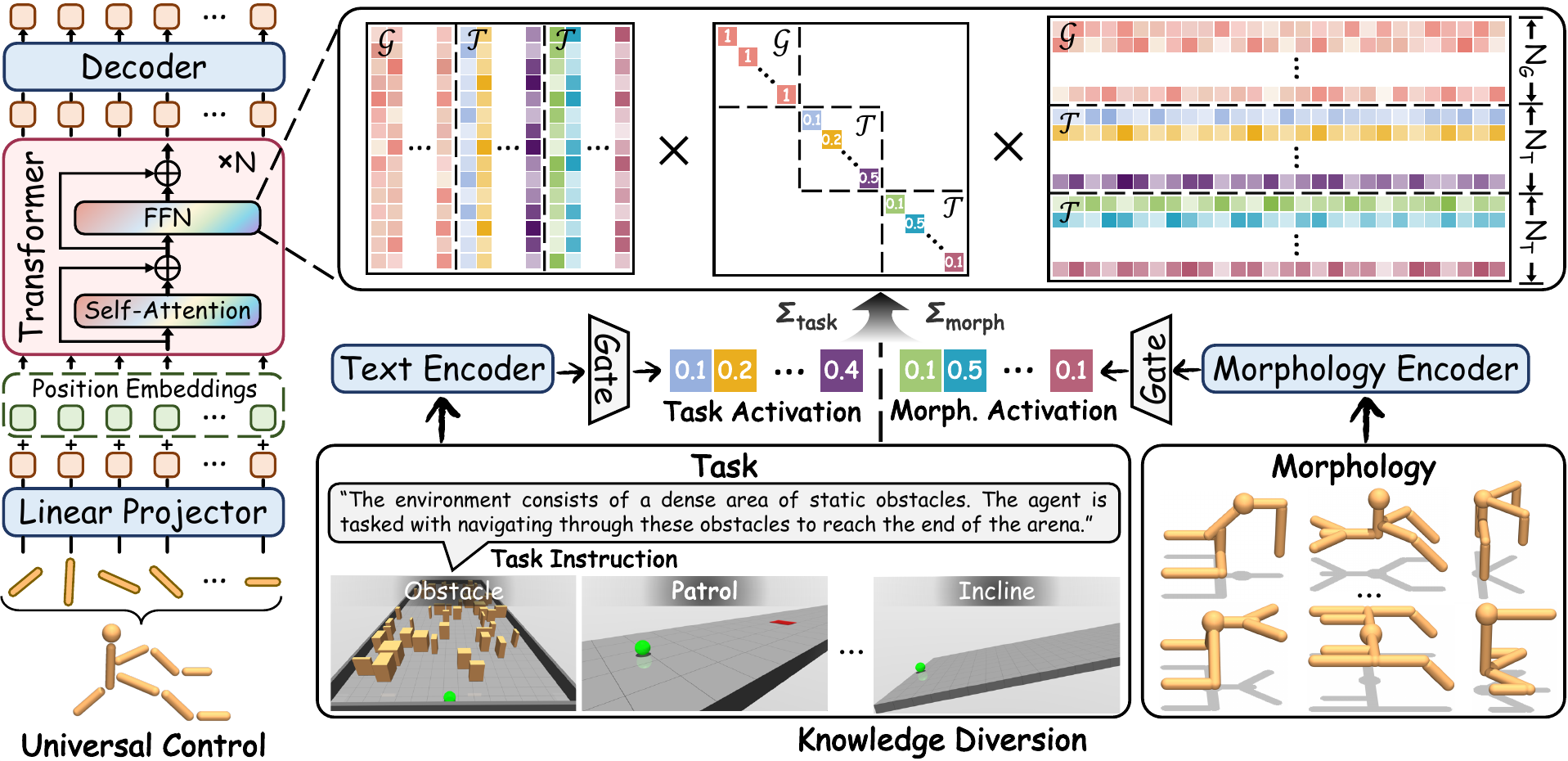}
    \caption{\textbf{Overview of DivMorph.} (a)~Morphology-Aware Transformer. It encodes the robot’s modular structure as a token sequence and models inter-limb interactions through a unified sequence representation, providing a generalizable controller across diverse morphologies. 
    (b)~Knowledge Diversion. Each randomly initialized weight matrix is factorized via SVD into a set of factor units, categorized into shared learngenes, morphology-specific tailors, and task-specific tailors. A dynamic soft gating mechanism selects the relevant tailors for each input while jointly updating the shared learngenes, enabling modular and disentangled representations across morphologies and tasks.}
    \label{fig:main}
    \vspace{-0.15in}
\end{figure*}

Transformers inherently capture interactions among variable-sized sets~\cite{lee2019set}, making them well suited for agents with diverse limb configurations. 
Leveraging this property, MetaMorph~\cite{guptametamorph} introduces the Morphology-Aware Transformer for universal control of robots with heterogeneous morphologies (Fig.~\ref{fig:main}a).

At each time step $t$, limb $i$ of robot $\kappa$ receives the concatenation of its proprioceptive observation $s^i_{\kappa,t}$ and morphology context $c^i_\kappa$, which is projected into a node embedding $\boldsymbol{e}_i$ via a shared linear layer. 
The set of node embeddings $\{\boldsymbol{e}_i\}_{i=1}^{N_\kappa}$ is processed by a transformer encoder to capture inter-limb interactions, where each layer comprises a multi-head self-attention (MSA) followed by a feed-forward network (FFN).

In the MSA module, each attention head computes query, key, and value vectors from the node embeddings via learnable linear projections: $\boldsymbol{q}_i=W_q \boldsymbol{e}_i$, $\boldsymbol{k}_i=W_k \boldsymbol{e}_i$ and $\boldsymbol{v}_i = W_v \boldsymbol{e}_i$, where $W_q, W_k, W_v \in \mathbb{R}^{D \times d}$ are trainable projection matrices, $D$ is the embedding dimension, and $d$ is the head dimension. 
Self-attention is then performed as
\begin{equation}
    A_i = \text{softmax}\Big(\frac{Q_i K_i^\top}{\sqrt{d}}\Big)V_i, \quad A_i \in \mathbb{R}^{N \times d},
\end{equation}
where $Q_i = [\boldsymbol{q}^i_1, \dots, \boldsymbol{q}^i_{N_k}]^\top$, $K_i = [\boldsymbol{k}^i_1, \dots, \boldsymbol{k}^i_{N_k}]^\top$, and $V_i = [\boldsymbol{v}^i_1, \dots, \boldsymbol{v}^i_{N_k}]^\top$.
For a multi-head attention module with $h$ heads, the outputs of all heads are concatenated and projected via a learnable matrix $W_o \in \mathbb{R}^{hd \times D}$:
\begin{equation}
    \text{MSA} = \text{concat}(A_1, \dots, A_h) W_o.
\end{equation}
The FFN consists of two linear transformations $W_\text{in}\in \mathbb{R}^{D\times D'}$ and $W_\text{out}\in \mathbb{R}^{D'\times D}$ with a GELU activation:
\begin{equation}
    \text{FFN}(x) = \text{GELU}(xW_\text{in} + b_1)W_\text{out} + b_2
\end{equation}
where $b_1$ and $b_2$ are the biases for the linear transformations.
% d: , and $D'$ denotes the hidden layer dimensions.

If global exteroceptive observations (e.g., a terrain height map or target goal) are available, they are encoded via a MLP and concatenated with the node embeddings to form enriched features. These features are then passed through a shared decoder $W_\text{dcd}$ to produce per-node actions, yielding the complete action vector for the robot at time $t$.

\section{DivMorph}

\subsection{Knowledge Decomposition in Weight Matrices}
To facilitate knowledge diversion for universal morphology control, we first decompose each weight matrix of the morphology-aware transformer into a set of factor units, providing a structural foundation for subsequent knowledge disentanglement and routing, as shown in Figure~\ref{fig:main}.

Formally, let $\theta = \{ W^{(1\sim L)}_q$, $W^{(1\sim L)}_k$, $W^{(1\sim L)}_v$, $W^{(1\sim L)}_o$, $W^{(1\sim L)}_{\text{in}}$, $W^{(1\sim L)}_{\text{out}}$, $W^{(1\sim L)}_{\text{dcd}}\}$\footnote{$W_{q}^{(1\thicksim L)}$ denotes the set $\{W_{q}^{(1)}, \dots, W_{q}^{(L)}\}$ for brevity. Similar notations throughout the paper follow this convention.}
denote the collection of weight matrices across the $L$ transformer layers, where $W^{(l)}_\star$ represents the weight matrix of type $\star \in \mathcal{S}$ in layer $l$, and $\mathcal{S} = \{q, k, v, o, \text{in}, \text{out}, \text{dcd}\}$.

Following the decomposition strategy in \cite{xie2025kind, xie2025divcontrol}, 
each matrix $W^{(l)}_\star$ is factorized via SVD:
\begin{equation}
    W_{\star}^{(l)} = U_{\star}^{(l)} \Sigma_{\star}^{(l)} {V_{\star}^{(l)}}^\top = \sum_{i=1}^{r} u_{\star}^{(l, i)} \sigma_{\star}^{(l, i)} v_{\star}^{(l, i)}
\label{equ:svd}
\end{equation}
where $r$ is the rank of $W_{\star}^{(l)}$. 
Each \mbox{rank-1} component $\Theta_{\star}^{(l, i)} = (u_{\star}^{(l, i)}, v_{\star}^{(l, i)})$ serves as a factor unit, and the singular values $\Sigma^{(l)}_\star=\{\sigma_{\star}^{(l, i)}\}_{i=1}^r$ modulate their contribution during reconstruction, which is controlled by dynamic soft gating mechanism (Sec.~\ref{sec:kdiv}). 

% During training, $U^{(l)}_\star$ is maintained orthogonal to preserve the disentangled structure of the knowledge factors.

% Each basic component can then be explicitly associated with a type of knowledge: \emph{morphology-agnostic} (shared), \emph{morphology-specific}, or \emph{task-specific}. The subsequent gating mechanism determines how these components are adaptively recombined for a given agent and task, enabling modular knowledge reuse and structured diversion.

\subsection{Knowledge Diversion across Morphology and Task}
\label{sec:kdiv}
To enable flexible adaptation across heterogeneous morphologies and tasks, the factor units are explicitly partitioned into \textit{morphology- and task-agnostic} learngenes
\begin{align*}
% \scalebox{0.92}{$
    \mathcal{G}^{(l)} = \left\{ \Theta_{\star}^{(l, i)} \mid i \in [0, N_G),\ \star \in \mathcal{S}\right\}
% $}
\end{align*}
and into \textit{morphology-specific} or \textit{task-specific} tailors 
\begin{align*}
    \mathcal{T}_{\kappa}^{(l)} &= \left\{ \Theta_{\star}^{(l, i)} \mid i \in [N_G, N_G + N_T^{\kappa}),\ \star \in \mathcal{S} \right\} \\
    \mathcal{T}_{\tau}^{(l)} &= \left\{ \Theta_{\star}^{(l, i)} \mid i \in [r - N_T^{\tau}, r),\ \star \in \mathcal{S} \right\}
\end{align*}
where the total components $r = N_G + N_T^{\kappa} + N_T^{\tau}$.

% \scalebox{0.92}{$
% \begin{aligned}
%     \mathcal{G} &= \{ \Theta_{\star}^{(l, i)} \mid i \in [0, N_G), \star \in \mathcal{S}, l \in [1,L] \} \\
%     \mathcal{T}_{\text{morph}} &= \{ \Theta_{\star}^{(l, i)} \mid i \in [N_G, N_G + N_T^{(\text{morph})}), \star \in \mathcal{S}, l \in [1,L] \} \\
%     \mathcal{T}_{\text{task}} &= \{ \Theta_{\star}^{(l, i)} \mid i \in [r - N_T^{(\text{task})}, r),\ \star \in \mathcal{S}, l \in [1,L] \}
% \end{aligned}$}

Unlike prior methods that employ binary gates conditioned on discrete classes~\cite{xie2025kind} or condition labels~\cite{xuctrlora}, we introduce a \textbf{soft gating mechanism} that models similarities across morphologies and tasks, thereby enhancing generalization to unseen morphologies and tasks.

Specifically, to enrich task semantics, each task is paired with a textual instruction $I_\tau$, which is encoded into a task embedding $e_\tau = E_\text{txt}(I_\tau)$ via a pretrained text encoder $E_\text{txt}$~\cite{wang2022text}. 
Similarly, each agent’s morphology, described by module and joint parameters $M_\kappa$ (e.g., limb radius, density, joint type, range), is encoded as $e_\kappa = E_\kappa(M_\kappa)$ by a dedicated morphology encoder $E_\kappa$.

The task and morphology embeddings are fed into task-aware and morphology-aware gates, $G_\tau$ and $G_\kappa$, producing soft weights over the corresponding tailor components:
% \begin{equation}
% \begin{aligned}
%     \boldsymbol{\sigma}^{\kappa}_T&= \mathrm{softmax}\Big(\mathrm{TopK}\big(G_\kappa(e_\kappa), k_\kappa\big)\Big)\\ \boldsymbol{\sigma}^{\tau}_T&= \mathrm{softmax}\Big(\mathrm{TopK}\big(G_\tau(e_\tau), k_\tau\big)\Big)
% \end{aligned}
% \end{equation}
\begin{equation}
    \boldsymbol{\sigma}^{\chi}_T= \mathrm{softmax}\Big(\mathrm{TopK}\big(G_\chi(e_\chi), k_\chi\big)\Big)\in \mathbb{R}^{N_T^{\chi}}, \ \chi \in \{\tau, \kappa\}
\label{equ:gate}
\end{equation}
% where $\boldsymbol{\sigma}_{\kappa}\in \mathbb{R}^{N_T^{\kappa}}$ and $\boldsymbol{\sigma}_{\tau}\in \mathbb{R}^{N_T^{\tau}}$.
where $\mathrm{TopK}(\cdot,k)$ applies a relevance-guided filtering that retains the $k$ largest gate scores before normalization.
Learngenes adopt unit scaling, i.e., $\boldsymbol{\sigma}_G=\mathbf{1}$.
The complete set of singular values $\Sigma=\{\boldsymbol{\sigma}_G, \boldsymbol{\sigma}^\kappa_T,  \boldsymbol{\sigma}^\tau_T\}$ thus determines the relative contribution of each factor unit during reconstruction.

In this way, DivMorph achieves an explicit separation of morphology specific and task specific knowledge by jointly optimizing learngenes $\mathcal{G}^{(l)}$, tailors $\mathcal{T}^{(l)}_{\{\kappa, \tau\}}$ and dynamic gates $G_{\{\kappa, \tau\}}$ via policy distillation, which in turn indirectly updates the reconstructed weight matrix $W^{(l)}$ in Eq.~\eqref{equ:svd} (see Algorithm~1 in App.~A).

% To maintain structural consistency in the factorized representation, w
We enforce an orthogonality constraint on the square factor (either $U$ or $V$) via a Cayley transform~\cite{trockmanorthogonalizing}, ensuring it remains on the orthogonal manifold~\cite{lezcano2019trivializations} and stabilizing the factorization during RL training (see App.~B for a brief proof).

% The final weight matrix is obtained via a gated combination of learngenes and tailors:
% \begin{equation}
%     \widetilde{W}^{(l)}_\star = \mathcal{G}^{(l)}_\star + \sum_{i=1}^{N_T^\kappa} \sigma^\kappa_i\mathcal{T}^{(l)}_{i,\star} + \sum_{i=1}^{N_T^\tau} \sigma^\tau_i \mathcal{T}^{(l)}_{i,\star}.
% \end{equation}

% During training, the dynamic gate activates relevant tailors according to the morphology and task embeddings, while shared learngenes are updated across all conditions. This design explicitly disentangles morphology- and task-agnostic knowledge from condition-specific knowledge, and all components—learngenes, tailors, and the dynamic gate—are jointly optimized end-to-end to facilitate modular reuse and zero-shot generalization.

\subsection{Efficient Deployment and Task Adaptation}
DivMorph trains a decomposable controller via knowledge diversion, enabling flexible composition across morphologies and tasks. Benefiting from the semantic generalization of the text and morphology encoders, the morphology- and task-aware gates acquire transferable routing behaviors, allowing the controller to \textit{zero-shot} adapt to unseen morphologies and semantically related tasks by selectively activating the corresponding tailors (Sec.~\ref{sec:trans_task} and Sec.~\ref{sec:trans_morph}).

Thus, for an agent with morphology $\kappa^\ast$ performing task $\tau^\ast$, whether previously seen or novel, its configuration $M_{\kappa^{\ast}}$ and task instruction $I_{\tau^\ast}$ are encoded and passed through the morphology- and task-aware gates ($G_\kappa$, $G_\tau$), which produce \textbf{\textit{sparse}} soft routing weights $\boldsymbol{\sigma}^\kappa_T$ and $\boldsymbol{\sigma}^\tau_T$ via TopK selection~(Eq.~\eqref{equ:gate}).
Leveraging this sparsity, \textbf{\textit{tailors not selected by the gates can be pruned}} for efficient deployment, yielding the effective weight:
\begin{equation}
    % \small
    \scalebox{0.92}{$
    \widetilde{W}^{(l)} = \mathcal{G}^{(l)} + \sum_{i=1}^{N_T^\kappa} \mathbf{1}_{[\sigma_{i}^\kappa > 0]} \sigma_{i}^\kappa \mathcal{T}^{(l)}_{\kappa,i} + \sum_{j=1}^{N_T^\tau} \mathbf{1}_{[\sigma_{j}^\tau > 0]}\sigma_{j}^\tau \mathcal{T}^{(l)}_{\tau,j}
    $}
\end{equation}
where $\mathbf{1}_{[\cdot]}$ is the indicator function selecting only non-zero entries of the sparse routing vectors $\boldsymbol{\sigma}^\kappa_T$ and $\boldsymbol{\sigma}^\tau_T$.

For tasks or morphologies with substantial shifts, the controller rapidly adapts with minimal computation by updating the selected tailors and applying minor adjustments to the learngenes, leveraging their condensed morphology- and task-agnostic knowledge, while the morphology- and task-aware gates remain fixed (see App.D.1 for details).

\section{Experimental Setup}
\paragraph{Environments}
We evaluate on the UNIMAL design space, using 100 training and 100 novel morphologies following MetaMorph~\cite{guptametamorph}. We consider 10 tasks with diverse objectives and scenarios~\cite{gupta2021embodied}, of which 5 serve as training tasks—Flat Terrain (FT), Variable Terrain (VT), Incline, Obstacle, and Patrol—for knowledge diversion, and 5 as novel tasks—Exploration, Escape, Point Navigation, Push Box Incline, and Manipulate Box—for evaluation. Task details are provided in Appendix~E.

\vspace{-0.1in}
\paragraph{Baselines}
We compare DivMorph with representative methods for morphology-aware control and cross-task transfer:
\textbf{1) MetaMorph}~\cite{guptametamorph} introduces the Morphology-Aware Transformer, the first transformer-based controller enabling universal control across a modular robot design space.
For a fair comparison, we additionally implement two policy-transfer variants of MetaMorph:
\textbf{2) MetaMorph}\arrowtrans directly transfers the policy from the most similar training task to the novel task.
\textbf{3) MetaMorph}\alambic distills policies from all training tasks into a single unified network for transfer.
\textbf{4) ModuMorph}~\cite{xiong2023universal} extends MetaMorph with morphology-conditioned fixed attention for improved morphology control.
\textbf{5) HyperDistill}~\cite{xiong2024distilling} distills ModuMorph’s knowledge into a hypernetwork that generates morphology-specific MLP controllers for efficient inference.
\textbf{6) GRL}~\cite{feng2023genes} condenses task-agnostic knowledge into inheritable neural fragments via population-based evolution, enabling effective knowledge transfer across tasks. 
% We adapt GRL for universal morphology control for generalization across diverse morphologies.

\vspace{-0.1in}
\paragraph{Ablations}
We conduct three ablations to assess the contributions of DivMorph’s core components.
\textbf{1) w/o KDiv.} We ablate knowledge diversion separately in the Transformer and decoder across morphologies and tasks to evaluate their individual contributions to policy adaptation.
\textbf{2) w/o Orthogonality.} We remove the orthogonality constraint on $U$ and $V$ to evaluate its impact on stabilizing RL training of the factorized parameters.
\textbf{3) w/o Multi-task Distillation.} Comparing MetaMorph and MetaMorph\alambic isolates the effect of multi-task distillation, evaluating the extent to which it facilitates policy transfer and generalization. Detailed results and analysis can be found in Sec.~\ref{sec:ablation}

\section{Results}
\begin{figure*}[!t]
    \centering
    \includegraphics[width=\linewidth]{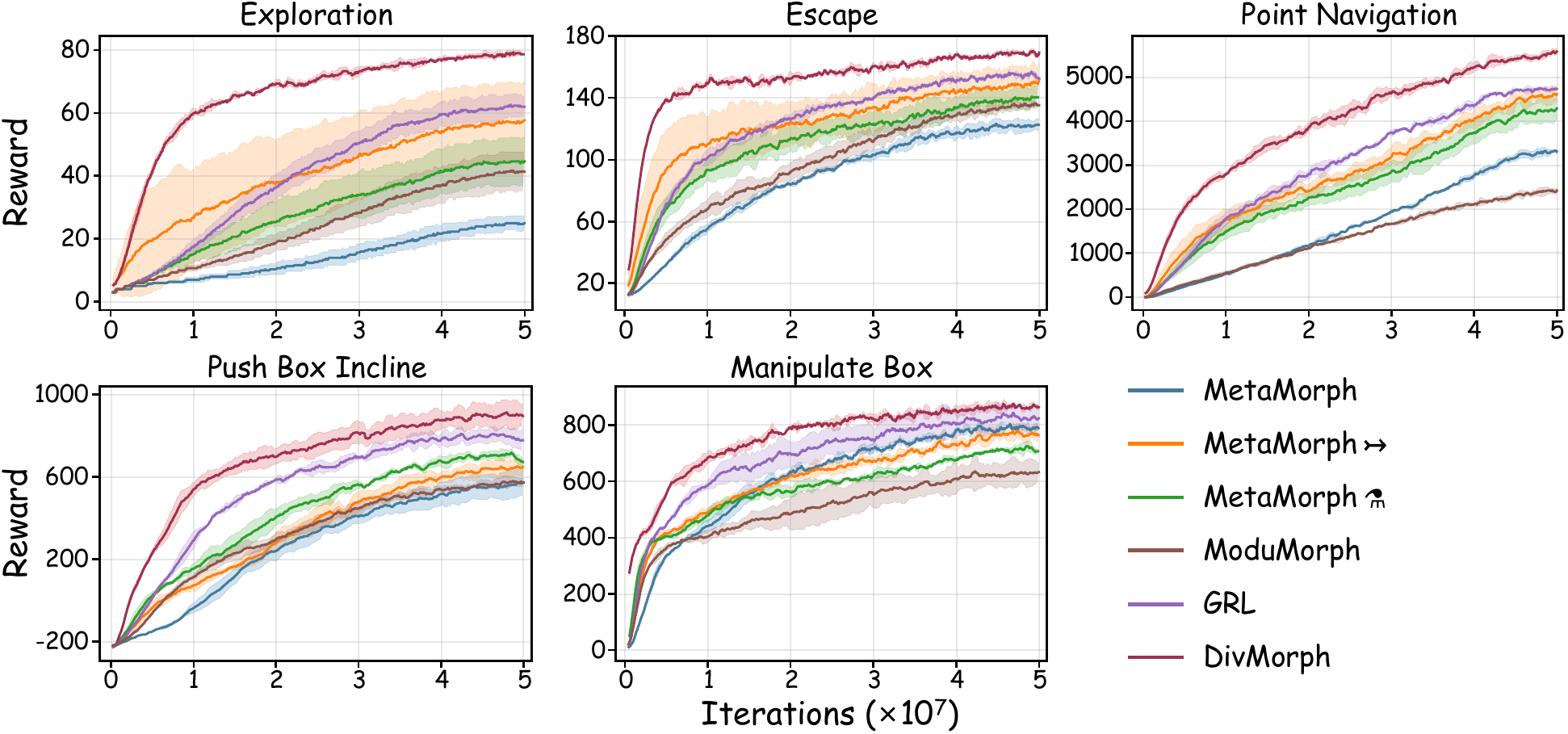}
    \vspace{-0.25in}
    \caption{\textbf{Comparison of policy transfer performance on novel tasks with training morphologies.} Training curves show the mean and standard deviation of rewards for 100 UNIMAL robots with \textit{training morphologies}, averaged over 3 runs per task. 
    DivMorph consistently outperforms baselines, demonstrating higher sample efficiency across all tasks.}
    \vspace{-0.15in}
    \label{fig:task_div_train}
\end{figure*}

\begin{figure*}[!t]
    \centering
    \includegraphics[width=\linewidth]{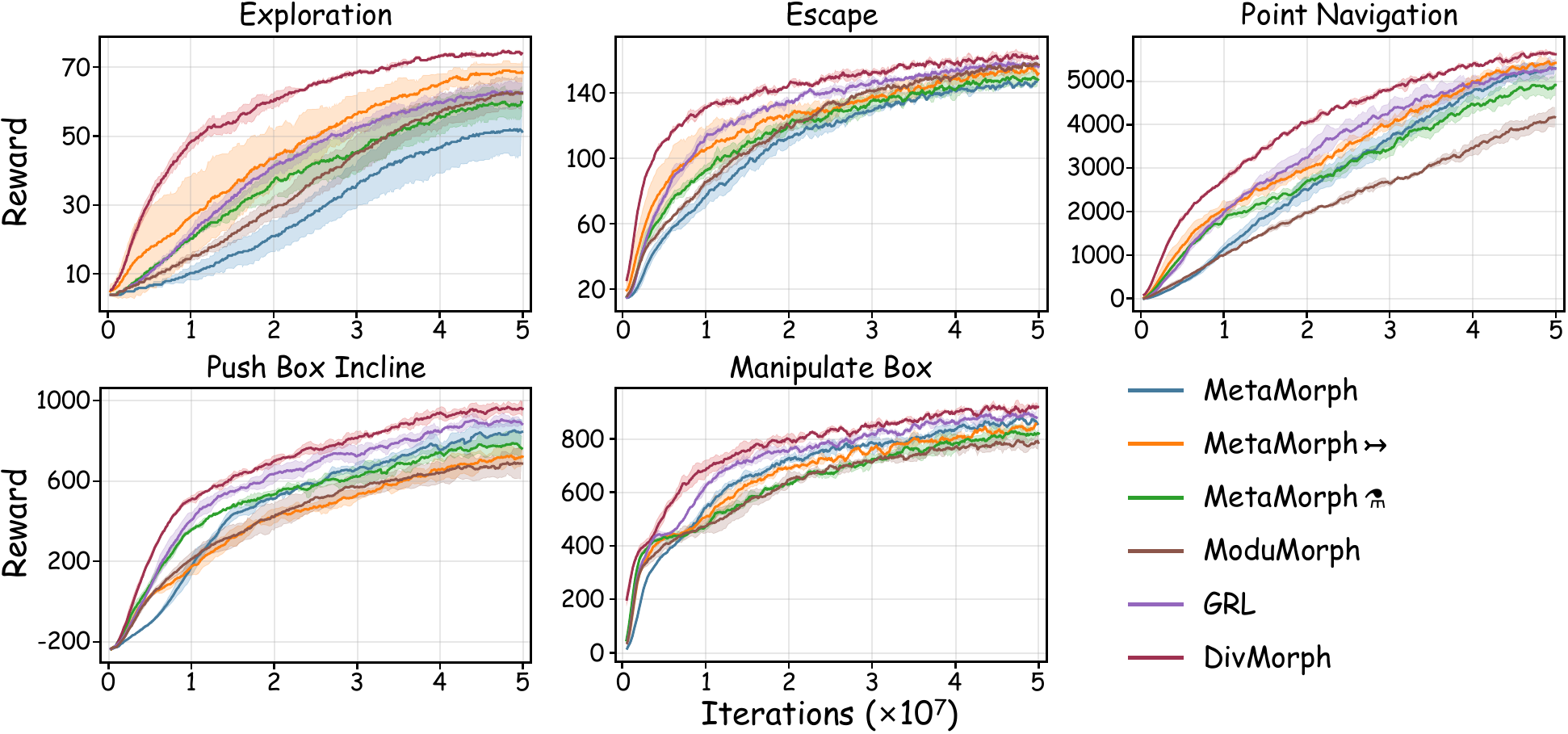}
    \vspace{-0.25in}
    \caption{\textbf{Comparison of policy transfer performance on novel tasks with novel morphologies.} Training curves show the mean and standard deviation of rewards for 100 UNIMAL robots with \textit{novel morphologies}, averaged over 3 runs per task. 
    DivMorph still maintains a clear advantage, further demonstrating its robust generalization and rapid adaptation to unseen morphologies and tasks.}
    \vspace{-0.15in}
    \label{fig:task_div_test}
\end{figure*}

\subsection{Effective Policy Transfer to Novel Tasks}
\label{sec:trans_task}
DivMorph factorizes the Morphology-Aware Transformer into shared learngenes and morphology- and task-specific tailors through knowledge diversion. 
To assess its effectiveness, we first evaluate policy transfer to novel tasks under \textit{universal control of training morphologies} (Fig.~\ref{fig:task_div_train}), isolating the contribution of task-level knowledge diversion.

Across all tasks, DivMorph achieves consistently superior performance, with clear improvements in both adaptation speed and final rewards. 
Sample efficiency improves by roughly $3\times$-$15\times$ to reach the performance of MetaMorph trained from scratch, with particularly pronounced gains in early training. 
Several tasks (e.g., Manipulate Box) even exhibit strong zero-shot performance, enabled by the semantic generalization of task instructions.

We observe that policy transfer generally surpasses training from scratch, but its effectiveness diminishes when the transferred priors are excessive or poorly aligned with the target task. 
MetaMorph\alambic underperforms MetaMorph\arrowtrans because aggregating multiple source policies introduces redundant and task-inconsistent priors, which in turn undermines generalization. 
Furthermore, both MetaMorph\alambic and MetaMorph\arrowtrans exhibit marked negative transfer on tasks with substantial behavioral mismatch (e.g., Manipulate Box), where inappropriate priors hinder efficient exploration.
In contrast, GRL attains competitive performance by transferring only task-agnostic fragments, retaining flexibility for new skills while avoiding irrelevant priors.

We further evaluate joint transfer to novel tasks and morphologies (Fig.~\ref{fig:task_div_test}), where naively reusing full policies (MetaMorph\alambic and MetaMorph\arrowtrans) leads to pronounced negative transfer and degraded performance. 
In contrast, GRL remains robust by transferring only task-agnostic fragments. Building on this principle, DivMorph combines task-agnostic learngenes with a task-aware gate that routes the most compatible tailors for each target task, mitigating negative transfer while enabling flexible and effective policy adaptation. 
These results highlight the critical role of diverting task-agnostic knowledge and recombining task-specific knowledge for scalable cross-task generalization.

\begin{figure*}[!t]
    \centering
    \includegraphics[width=\linewidth]{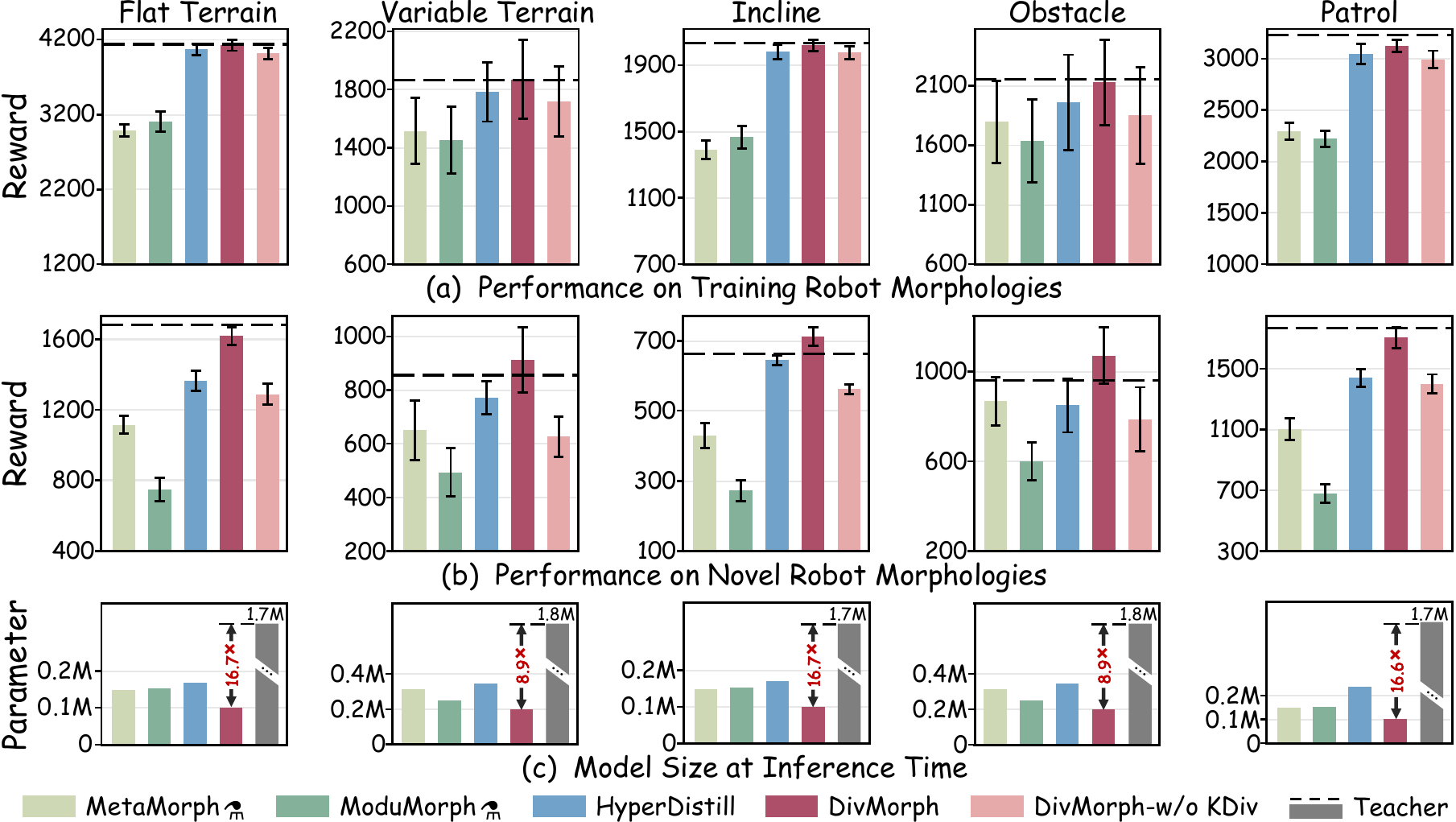}
    \vspace{-0.2in}
    \caption{\textbf{Comparison of single-agent deployment performance on training and novel morphologies.} 
    Bars show the mean and standard deviation of rewards for 100 UNIMAL robots, averaged over 3 runs per task (\textbf{a}, \textbf{b}), together with the corresponding model parameter count (\textbf{c}). 
    DivMorph attains substantial model compression while matching the teacher models (MetaMorph pre-trained policies) on training morphologies, which approximate an upper performance bound.
    On novel morphologies, DivMorph exhibits strong zero-shot generalization, at times even exceeding the teacher.}
    \vspace{-0.15in}
    \label{fig:morph_div}
\end{figure*}

\subsection{Efficient Deployment for Morphology Control}
\label{sec:trans_morph}
DivMorph separates task- and morphology-level knowledge into independent learngenes and tailors, enabling a policy that can be flexibly recombined at deployment. This modular design allows controllers to adapt to diverse target morphologies and tasks by activating only relevant components while pruning irrelevant ones, substantially reducing the deployed model size (Fig.~\ref{fig:morph_div}).

For training morphologies, DivMorph matches teacher performance—approaching an upper bound—while reducing model size by 8.9$\times$ and 16.7$\times$ (Fig.~\ref{fig:morph_div}a,c).
This demonstrates that the continuous dynamic gate provides sufficient representational capacity for DivMorph to train decomposable transformers that achieve effective knowledge diversion and high-fidelity morphology encoding.

Importantly, the morphology encoder and morphology-aware gate together facilitate effective cross-morphology generalization, enabling DivMorph to maintain superior performance on unseen agents and even exceed the transferred teacher policy on several tasks (Fig.~\ref{fig:morph_div}b).
In contrast, HyperDistill, despite strong compression on training morphologies, generalizes poorly to novel ones; its hypernetwork, which generates full MLP parameters, is prone to overfitting, limiting adaptability under distribution shifts.

Ablations on morphology knowledge diversion further highlight the importance of separating morphology-agnostic and morphology-specific components for efficient adaptation and generalization.

\begin{figure*}[!t]
    \centering
    \includegraphics[width=\linewidth]{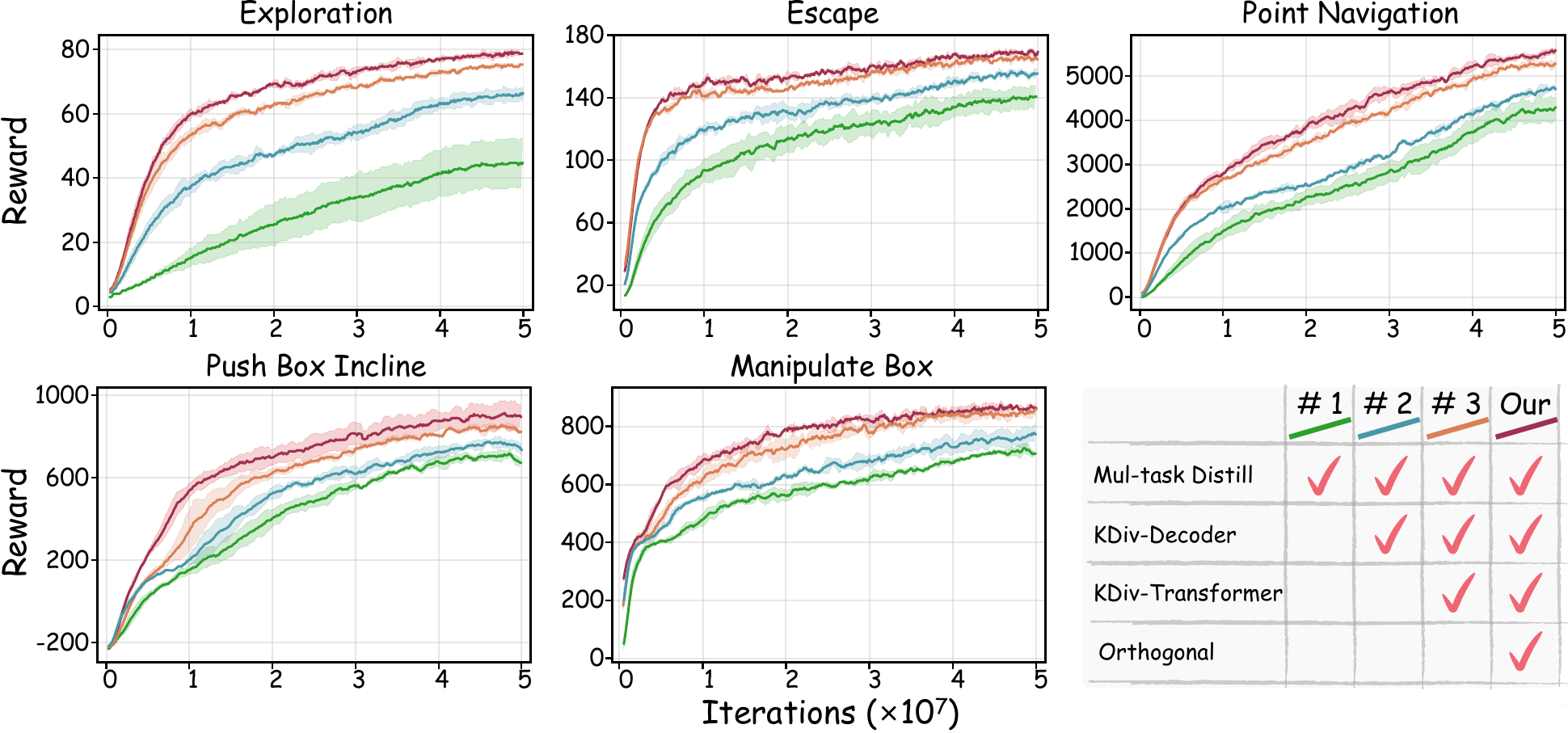}
    \vspace{-0.2in}
    \caption{\textbf{Ablation study on DivMorph.} Training curves comparing DivMorph with variants that remove transformer-level diversion, decoder-level diversion, or orthogonality constraints, highlighting their contributions to learning efficiency, stability, and final performance.}
    \vspace{-0.2in}
    \label{fig:abl}
\end{figure*}

\begin{figure}[!t]
    \centering
    \includegraphics[width=\linewidth]{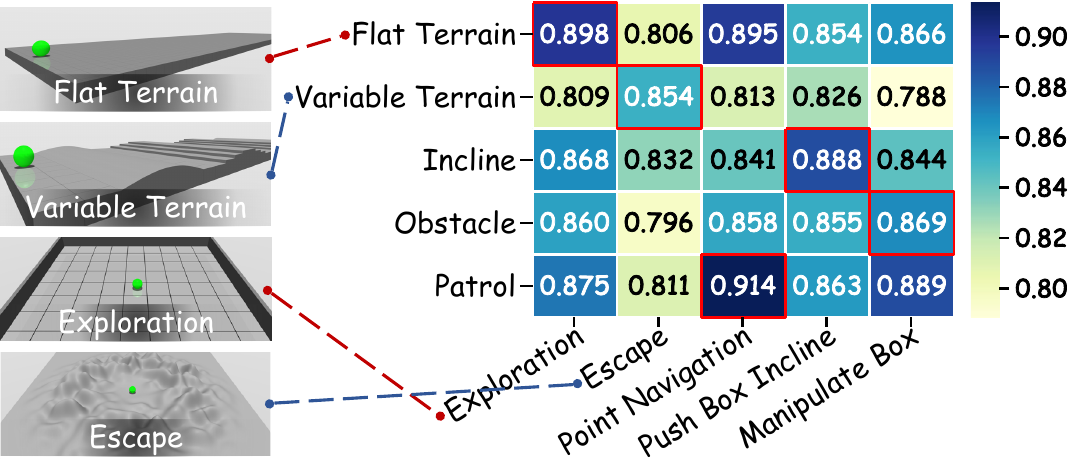}
    \vspace{-0.2in}
    \caption{\textbf{Visualization of task-aware gate activations.} Correlation matrix of training and unseen tasks, illustrating inter-task similarity and task-specific routing patterns.}
    \vspace{-0.2in}
    \label{fig:attn_task}
\end{figure}

\subsection{Ablation and Analysis}
\label{sec:ablation}
\subsubsection{Effect of Knowledge Diversion}
Ablation of morphology-level knowledge diversion was presented in Sec.~\ref{sec:trans_morph}. Here, we perform a detailed ablation of task-level knowledge diversion (Fig.~\ref{fig:abl}). Diverting knowledge in both the transformer backbone and decoder is crucial for extracting task-agnostic representations; removing either component slows convergence and lowers final returns.
Specifically, Transformer-level diversion captures generalizable task patterns, accelerating early learning, and decoder-level diversion encodes fine-grained adaptations, improving precise control and transfer, with both mechanisms together enabling scalable cross-task generalization.

\subsubsection{Effect of Orthogonality Constraints}
Removing the orthogonalization constraint on $U$ or $V$ (each being one of the square factors) degrades performance and destabilizes training (Fig.~\ref{fig:abl}). Orthogonal matrices lie on a compact manifold, whereas non-orthogonal ones introduce unbounded, unstable degrees of freedom. Constraining the square factor, which governs the isometric properties of the full mapping, eliminates these unstable directions, stabilizing RL training and ensuring well-conditioned knowledge factors (a simple proof is provided in App.~B).

\subsubsection{Effect of Morphology- \& Task-aware Gate}
\begin{figure}[!t]
    \includegraphics[width=\linewidth]{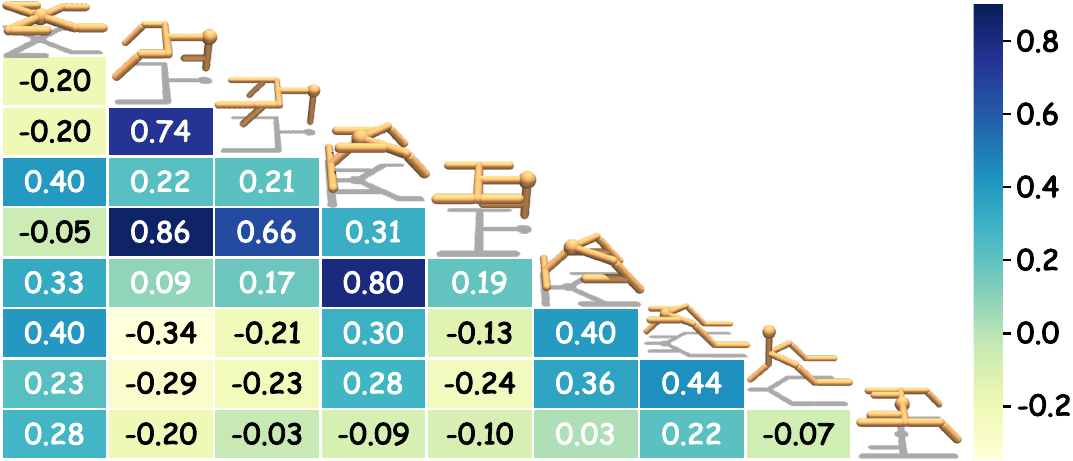}
    \vspace{-0.2in}
    \caption{\textbf{Visualization of morphology-aware gate activations.} Correlation matrix of diverse robot morphologies, illustrating morphology-speific routing and learned generalization.}
    \label{fig:attn_morph}
    \vspace{-0.2in}
\end{figure}

To better understand DivMorph’s adaptive control mechanism, we visualize the morphology- and task-aware gates in Fig.~\ref{fig:attn_task} and Fig.~\ref{fig:attn_morph}, respectively. 

The task-aware gate exhibits a clear semantic organization. Tasks with similar objectives activate highly overlapping subsets of tailors (e.g., Variable Terrain and Escape), indicating that textual instructions effectively complement and enrich the underlying task semantics. Together with the pre-trained text encoder, this gating mechanism provides a strong inductive bias for cross-task generalization.

Similarly, the morphology-aware gate models morphologies through topological and parametric features, generating consistent activations for structurally similar robots while distinguishing markedly different ones, thereby capturing shared regularities rather than memorizing individual designs and enabling reliable transfer across morphologies.

\section{Conclusion}
We introduce DivMorph, a novel framework for universal morphology control that performs knowledge diversion across both morphology and tasks to construct a modular controller.
By disentangling the transformer into shared learngenes and morphology- and task-specific tailors, DivMorph enables flexible recomposition at deployment through selective activation of relevant components.
This facilitates efficient policy deployment while enabling effective transfer to novel tasks and robust generalization to unseen morphologies.
Extensive experiments show that DivMorph consistently outperforms existing methods, achieving significant improvements in sample efficiency and overall performance while substantially reducing model size.

\section*{Acknowledgements}
We sincerely  appreciate Freepik for contributing to the figure design. This research was supported by the Jiangsu Science Foundation (BG2024036, BK20243012), the National Natural Science Foundation of China (62125602, U24A20324, 92464301, 62306073), China Postdoctoral Science Foundation (2022M720028, 2025T180432), the Xplorer Prize, and the Fundamental Research Funds for the Central Universities (2242025K30024).

% \section*{Impact Statement}

% \clearpage
\bibliography{icml2026}
\bibliographystyle{icml2026}

\end{document}